\documentclass[journal]{IEEEtran}      
\usepackage{graphics} 
\usepackage{epsfig} 
\usepackage{mathptmx} 
\usepackage{times} 
\usepackage{amsmath} 
\usepackage{amssymb}  
\usepackage{cite}
\usepackage{array}
\usepackage{subfigure}
\usepackage{color, soul}
\usepackage{algorithmic,algorithm}
\usepackage{multirow}
\usepackage{hyperref}

\usepackage[utf8]{inputenc}
\usepackage{color,soul}
\usepackage[T1]{fontenc}
\newtheorem{lemma}{Lemma}
\newtheorem{remark}{Remark}

\graphicspath{{figs/}}

\title{
Receding Horizon Motion Planning for\\
Multi-Agent Systems: A Velocity Obstacle Based Probabilistic Method
}

\author{
	Xiaoxue Zhang, 
	Jun Ma,	
		Zilong Cheng,
	Sunan Huang,
		and Tong Heng Lee	
	\thanks{X. Zhang, Z. Cheng, and T. H. Lee are with the NUS Graduate School for Integrative Sciences and Engineering, National University of Singapore, Singapore 119077 (e-mail: xiaoxuezhang@u.nus.edu; zilongcheng@u.nus.edu; eleleeth@nus.edu.sg).}
	\thanks{J. Ma is with the Department of Mechanical Engineering, University of California, Berkeley, CA 94720 USA (e-mail: jun.ma@berkeley.edu).}
	\thanks{S. Huang is with the Temasek Laboratories, National University of Singapore, Singapore, 117411 (e-mail: tslhs@nus.edu.sg).}	
	\thanks{This work has been submitted to the IEEE for possible publication.
		Copyright may be transferred without notice, after which this version may
		no longer be accessible}
}

\begin{document}

\maketitle

\begin{abstract}
In this paper, a novel and innovative methodology for feasible motion planning in the multi-agent system is developed. On the basis of velocity obstacles characteristics, the chance constraints are formulated in the receding horizon control (RHC) problem, and geometric information of collision cones is used to generate the feasible regions of velocities for the host agent. By this approach, the motion planning is conducted at the velocity level instead of the position level. Thus, it guarantees a safer collision-free trajectory for the multi-agent system, especially for the systems with high-speed moving agents. Moreover, a probability threshold of potential collisions can be satisfied during the motion planning process. In order to validate the effectiveness of the methodology, different scenarios for multiple agents are investigated, and the simulation results clearly show that the proposed approach can effectively avoid potential collisions with a collision probability less than a specific threshold.
\end{abstract}


\section{Introduction}
Motion planning is one of the essential components of intelligent robots, and an efficient trajectory can effectively improve robots' intelligence and autonomy~\cite{gu2015tunable,hang2021path}. The velocity obstacles approach is a geometry-based method that defines conic regions as constraints on the feasible velocities for agents. The concepts of collision cones and velocity obstacles approach are introduced in~\cite{fiorini1998motion}. This kind of method can be used to find the range of feasible velocity quickly with minimal obstacle information without any prior knowledge or prediction, after given the collision scenario geometrically~\cite{douthwaite2019velocity}. There are several variants of the velocity obstacles approach, i.e., reciprocal velocity obstacles~\cite{van2008reciprocal}, accelerated velocity obstacles~\cite{van2011reciprocal}, generalized velocity obstacles~\cite{wilkie2009generalized}, hybrid reciprocal velocity obstacles~\cite{snape2011hybrid}, etc. However, most of these variants consider the feasible velocity direction with neglecting the optimality of the planned trajectory.

In order to find a feasible velocity based on the collision region provided by velocity obstacles, receding horizon control (RHC), which is also known as model predictive control (MPC), can be utilized to find the optimal velocity and position~\cite{ji2016path}. It is one typical effective approach that has been widely applied in the decision making, planning, and control of robots~\cite{zhang2021sequential,cheng2021admm,zhang2020accelerated}. MPC has the capability of addressing various constraints as part of the control synthesis problem~\cite{defoort2011decentralized, cheng2017decentralized, zhang2019integrated, luis2019trajectory}. In terms of the uncertainties of path planning, one can characterize these uncertainties in a probabilistic manner and find the optimal sequence of control inputs subject to the chance constraints, which means that the probability of collision avoidance should not be lower than a user-defined threshold~\cite{zhu2019chance,da2019collision,ma2020data}. Such a probabilistic approach has multiple advantages compared to the traditional box constraints approach, as most of the uncertainties can be represented by a stochastic model instead of a bounded set~\cite{ma2021optimal}. Besides, the probability threshold will influence the conservatism of the planned trajectory. A proper probability threshold can balance conservatism and performance. There are a number of previous works regarding the chance constrained path planning with the existence of obstacles~\cite{blackmore2006probabilistic, zhang2020trajectory, blackmore2011chance, lenz2015stochastic}.  However, most of the existing probabilistic approaches still stay on the position-based level to achieve motion planning, which means the position information is mainly utilized to make a decision. Thus, these approaches at the position-based level could not be suitable even valid in some scenes, especially in fast-moving scenarios. 

This paper presents a novel approach to plan the optimal trajectories for the multi-agent system through the confinement of collision probability, based on the information of velocity obstacles. This approach generates a feasible region of velocity for the host agent when computing an optimal trajectory. Then the feasible region is further formulated as probabilistic collision constraints. This method is capable of planning the trajectories at the velocity level, which is much more meaningful than the traditional counterpart at the position level, especially effective for the scenarios with high-speed obstacles. Also, a collision probability can be confined within a specific range in the proposed approach. The structure of this paper is as follows. In Section~\ref{section:vo intro}, the basic concept and definitions of the velocity obstacles method are introduced. Section~\ref{section:problem_ori} formulates the chance constrained receding horizon optimization problem based on the velocity obstacles. In Section~\ref{section:vo_ccProblem}, we transform the probabilistic constraints in the model predictive control (MPC) problem into deterministic constraints and then solve this problem by multiple shooting method. Section~\ref{section:results} shows the simulation results in different scenarios, and conclusion is presented in Section~\ref{section:conclusion}.

\section{Velocity Obstacles}
\label{section:vo intro}
\begin{figure}[thpb]
	\centering
	\includegraphics[width=0.45\textwidth, trim=80 130 80 0,clip]{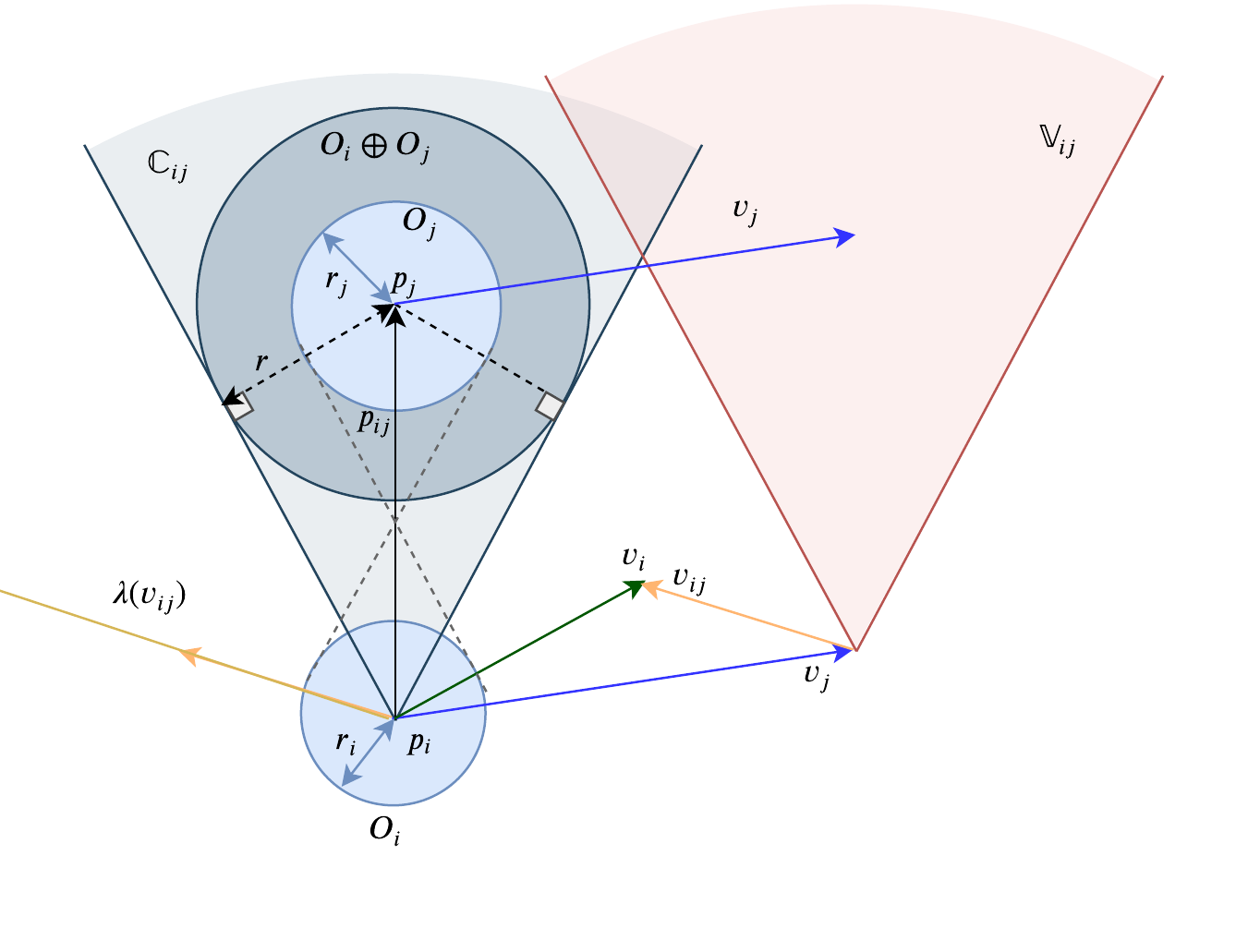}
	\caption{Velocity obstacles}
	\label{fig:vo_brief}
\end{figure}

In this section, the basic concept of the velocity obstacles approach is introduced, and the geometric illustration is shown in Fig.~\ref{fig:vo_brief}. The related parameters and variables in this figure are explained below.

Let all circular moving objects $O_i$ and $O_j$ be centered at $p_i \in \mathbb R^{n}$ and $p_j \in \mathbb R^{n}$ with radius $r_i$ and $r_j$ and velocities $v_i \in \mathbb R^{n}$ and $v_j\in \mathbb R^{n}$, respectively, where 
\begin{IEEEeqnarray}{rCl}
	O_i &=& \{p_i+\mu r_i\;|\; \|\mu\|_2 \leq 1\}\nonumber
	\\ O_j &=& \{p_j+\mu r_j\;|\; \|\mu\|_2 \leq 1\}.
\end{IEEEeqnarray}
Let $\oplus$ denotes the Minkowski sum operation, and we have
\begin{IEEEeqnarray}{rCl}
	O_i \oplus O_j  & = & \{m_i+m_j \;|\; m_i\in O_i, m_j\in O_j\}.
\end{IEEEeqnarray}
The relative velocity between $i$th object and $j$th object is $v_{ij}:=v_i-v_j$. Let $\lambda(v_{ij})$ denote the ray with direction $v_{ij}$ from the position $p_i$, where
\begin{IEEEeqnarray}{rCl}
	\lambda(v_{ij}) & = & \{p_i + \lambda v_{ij} \;|\; \lambda \geq 0 \}.
\end{IEEEeqnarray}
Then the object $i$ and the object $j$ will collide if and only if 
\begin{IEEEeqnarray}{rCl}
	\lambda(v_{ij}) \cap \left(O_i \oplus O_j\right) \neq \emptyset.
\end{IEEEeqnarray}
Therefore, a collision cone $\mathbb C_{ij}$ can be represented by
\begin{IEEEeqnarray}{rCl}
	\mathbb C_{ij} &=& \left\{ v_{ij}\;\left|\; \left(O_i \oplus O_j\right)  \cap \lambda_{ij}(v_{ij}) \neq \emptyset \right.\right\}.
\end{IEEEeqnarray}
In order to determine whether the velocity $v_i$ of the $i$th object has a risk of collision with the $j$th object, the velocity obstacle $\mathbb V_{ij}$ is defined as 
\begin{IEEEeqnarray}{rCl}
	\mathbb V_{ij} = \left\{ v_i \;|\; (v_i-v_j) \in \mathbb C_{ij} \right\},
\end{IEEEeqnarray}
which is equivalent to
\begin{IEEEeqnarray}{rCl}
	\mathbb V_{ij} = v_j + \mathbb C_{ij},
\end{IEEEeqnarray}
for the $i$th object. Any velocity $v_i \in \mathbb V_{ij}$ will result in a collision, as shown in Fig.~\ref{fig:vo_brief}.

The set of all moving surrounding objects can be considered as obstacles for the host object $i$. Assume the set of all moving objects is $\mathbb N_n = \{1,2,\cdots, n\}$. The composite velocity obstacles and collision cones are 
\begin{IEEEeqnarray}{rCl}
	\mathbb V_i &=& \cup_{j\neq i} \mathbb V_{ij} \nonumber \\
	\mathbb C_i &=& \cup_{j\neq i} \mathbb C_{ij}.
\end{IEEEeqnarray}

\section{Problem Statement}
\label{section:problem_ori}
\subsection{Objects Model and Collision Chance Constraints} 
\subsubsection{Objects Model}
The dynamics of each planar object $i \in \mathbb N_n$ can be represented by any stochastic nonlinear or linear, continuous-time or discrete-time model, where
\begin{IEEEeqnarray}{rCl}
\label{eq:dynamics}
	\dot x_i = f_i(x_i, u_i) + \omega_i\quad\text{or}\quad x_i^{k+1} = g_i\left(x_i^k, u_i^k\right) + \omega_i^k,
\end{IEEEeqnarray}
where $x_i = \begin{bmatrix} p_i^T, v_i^T \end{bmatrix}^T \in \mathcal X_i \subset \mathbb R^{n_x}$ denotes the state vector consisting of positions and velocities, $u_i\in \mathcal U_i \subset \mathbb R^{n_u}$ is the control input vector, $n_x$ and $n_u$ represent the dimension of the state vector and the control input vector, respectively, and $\mathcal X_i$ and $\mathcal U_i$ are the state and control space of the $i$th object, respectively. In the discrete-time model, $k$ means the $k$th time step for the objects. $f_i$ and $g_i$ are the nonlinear  continuous-time and discrete-time dynamics models of the $i$th object. We consider the Gaussian process noise of velocity in the objects model, i.e., $\omega_i \sim \mathcal N(0,W_i)$ with a diagonal covariance matrix $W_i$. In the following, we use the discrete-time dynamics model to illustrate our approach.

\subsubsection{Collision Chance Constraints}
According to the previous introduction of velocity obstacles, we can obtain the collision condition of the object $i$ with respect to the object $j$, which is defined as 
\begin{IEEEeqnarray}{rCl}
	v_i^k \notin \mathbb{V}_i \quad\text{or}\quad v_{ij}^k \notin \mathbb{C}_i .
\end{IEEEeqnarray}
Since there is additional noise on the velocity of objects, the velocity of each object can be described as random variables $v_i^k:=\hat v_i^k + \omega_i^k \sim \mathcal N(\hat v_i^k, W_i)$, where $\hat v_i^k$ is the mean of the velocity and $\omega_i^k$ is the additional noise at time $k$. Hence, the collision avoidance constraints can be described in a probabilistic manner, where for each object $i$, the chance constraints can be expressed as
\begin{IEEEeqnarray}{rrCl}
	&\operatorname{Pr}\left(v_i^k \notin \mathbb V_{i}^k\right) &\ge& 1- \delta_i, \quad \forall i\in \mathbb N_n\IEEEyesnumber\IEEEyessubnumber\\
	\text{or}\quad& 	\operatorname{Pr}\left(v_{ij}^k \notin \mathbb C_{i}^k\right) &\ge& 1- \delta_i, \quad \forall i\in \mathbb N_n,\IEEEyessubnumber\label{eq:condition}
\end{IEEEeqnarray}
where $\delta_i$ is the probability threshold of the collision risk. 

\subsection{Distributed Collision Avoidance Problem}
Here, we formulate a distributed collision avoidance problem. For each object $i \in \mathbb N_n$, we formulate a discrete-time chance constrained optimization problem on $N$ prediction steps with a sampling time $\Delta t$. 

\textit{Problem 1: (Optimization with Probabilistic Chance Constraints)} For each host object $i$, the position of the other objects $p_j, \forall j \in \mathbb N_n, j\neq i$, the initial state $\hat x^0_i$ with uncertain noise, the reference state vector $x_{\text{ref},i}$ and the collision probability threshold $\delta_i$ have been provided. The objective is to compute the optimal trajectories and control inputs for all of the objects. Thus, these objects can move from their initial states to the target states while maintaining the collision probability below the given threshold. The problem is defined as
\begin{IEEEeqnarray*}{rCl}
	\min\limits_{x_i^{1:N}, u_i^{0:N-1}} & \quad & \sum_{k=0}^{N-1} \left\|x_i^k-x_{\text{ref},i}^k \right\|_{Q_{i}} + \left\|u_i^k\right\|_{R_{i}} \\
	\operatorname{subject\ to} & \quad & x_i^{k+1} = g_i\left(x_i^k, u_i^k\right) + \omega_i^k \\
	&\quad& v_i^{k+1} \sim \mathcal N\left(\hat v_i^{k+1}, W_i\right) \\
	&\quad& \operatorname{Pr}\left(v_i^{k+1} \notin \mathbb V_{i}^{k+1}\right) \ge 1- \delta_i, \quad \forall i\in \mathbb N_n \\
	&\quad& \underline x_i^{k+1} \leq x_i^{k+1} \leq \overline x_i^{k+1} \\
	&\quad& \underline u_i^{k} \leq u_i^{k} \leq \overline u_i^{k} \\
	&\quad& x_i^{k+1}\in \mathcal X_i, \quad u_i^{k} \in \mathcal U_i, \yesnumber\label{eq:ori_problem}
\end{IEEEeqnarray*}
where $\|x_i^k-x_{\text{ref},i}^k \|_{Q_{i}} = \langle (x_i^k-x_{\text{ref},i}^k),\; Q_i(x_i^k-x_{\text{ref},i}^k) \rangle, \|u_i^k\|_{R_{i}}= \langle u_i^k, R_iu_i^k \rangle$, $Q_i$ and $R_i$ are weighting matrices to penalize the deviation from the reference states and the unnecessary large control inputs, respectively. $\underline x_i,\overline x_i,\underline u_i, \overline u_i$ are the lower bound and upper bound of state variables and control inputs.

\begin{remark}
	In the problem formulation~\eqref{eq:ori_problem}, the collision avoidance constraints are based on the $(k+1)$th time step. But in the next mathematical transformation, our discussion is based on the $k$th time step for simplicity.
\end{remark} 

\section{Receding Horizon Optimization Problem with Chance Constraints}
\label{section:vo_ccProblem}
\begin{figure}[thpb]
	\centering
	\includegraphics[width=0.5\textwidth, trim=80 150 80 0,clip]{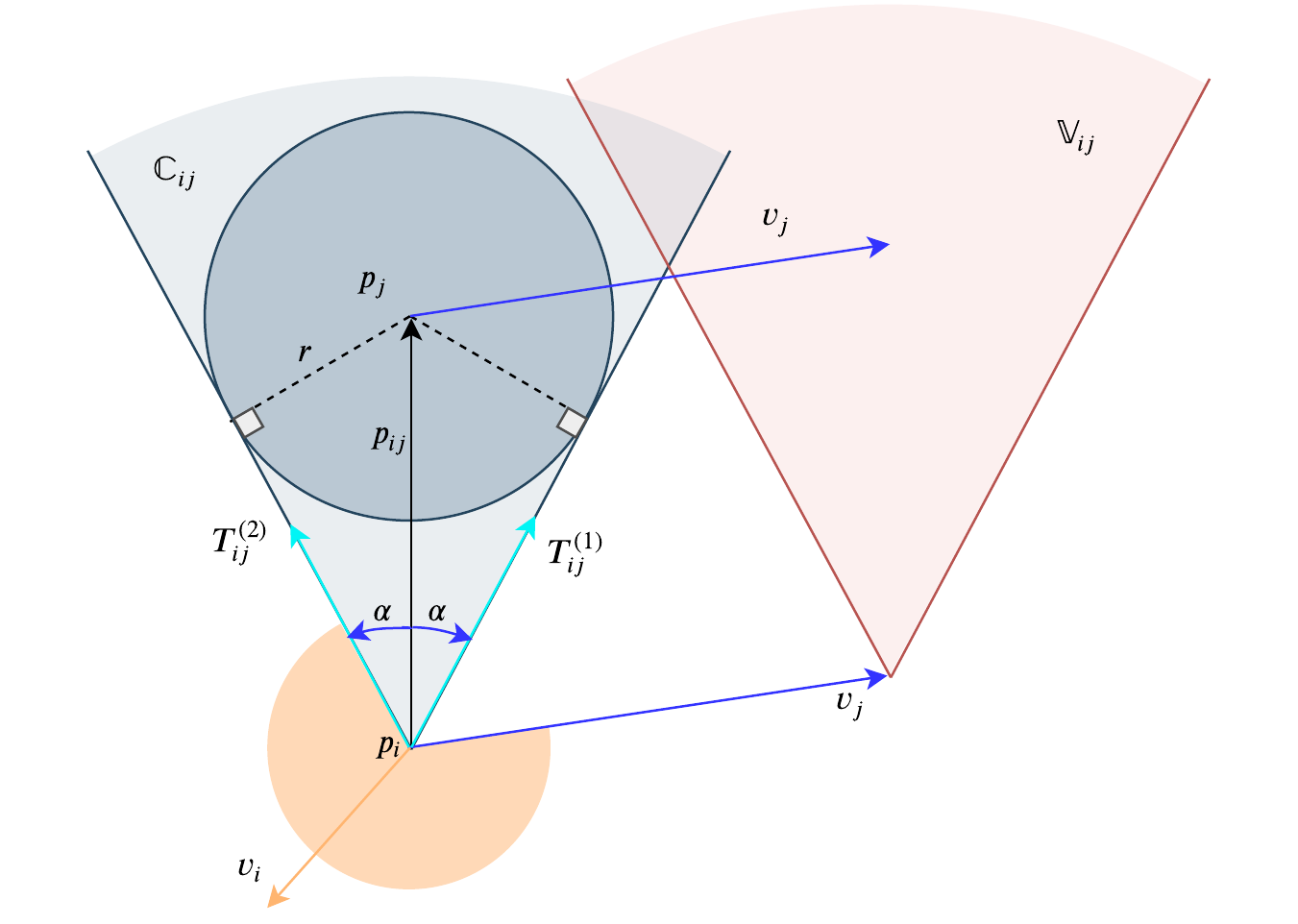}
	\caption{Feasible region analysis based on the velocity obstacle $\mathbb V_{ij}$}
	\label{fig:vo_ori}
\end{figure}

The chance constraints of velocity obstacles $\operatorname{Pr}(v_i^k \notin \mathbb V_{i}^k) \ge 1- \delta_i, \forall i\in \mathbb N_n$ mean that the risk of collision should be less than $\delta_i$; however, it is hard to determine it. Therefore, we need to do some mathematical manipulations to make it much easier to solve. The composite collision cone $\mathbb C_i^k$ and velocity obstacles $\mathbb V_i^k$ are the union of $\mathbb C_{ij}^k$ and $\mathbb V_{ij}^k$ with respect to object $j$ with $j\neq i$. Thus, we start to analyze it in terms of one collision cone $\mathbb C_{ij}^k$ or one velocity obstacle $\mathbb V_{ij}^k$. All of the pertinent variables and their relationships are demonstrated in Fig.~\ref{fig:vo_ori}. 

Define the vector $p_{ij}^k:=p_j^k-p_i^k \in \mathbb R^{n_p}$, where $n_p$ is the dimension of the position vector, and the radius of the circle is given by $r=r_i+r_j$. Then, the angle between $p_{ij}^k$ and the boundary line of the collision cone is the angle $\alpha_{ij}^k$, as shown in Fig.~\ref{fig:vo_ori}. Obviously, $\sin\alpha_{ij}^k= \frac{r}{|p_{ij}^k|}$, and $\cos \alpha_{ij}^k=\sqrt{1- \sin^2 \alpha_{ij}^k} = \frac{\sqrt{|p_{ij}^k|^2-r^2}}{|p_{ij}^k|}$. Based on the rotation relationship of the vector $p_{ij}^k$, we can obtain the two tangent directional vectors $T_{ij,1}^{k}$ and $T_{ij,2}^{k}$, where
\begin{IEEEeqnarray*}{rCl}
	T_{ij,1}^{k} & = & \begin{bmatrix} \cos\alpha_{ij}^k & \sin\alpha_{ij}^k \\ -\sin\alpha_{ij}^k & \cos\alpha_{ij}^k \end{bmatrix} p_{ij}^k \\
	T_{ij,2}^{k} & = & \begin{bmatrix} \cos\alpha_{ij}^k & -\sin\alpha_{ij}^k \\ \sin\alpha_{ij}^k & \cos\alpha_{ij}^k \end{bmatrix} p_{ij}^k. \yesnumber
\end{IEEEeqnarray*}

\subsection{Analysis on Velocity Obstacles}
\begin{figure}[thpb]
	\centering
	\includegraphics[width=0.35\textwidth, trim=80 130 80 0,clip]{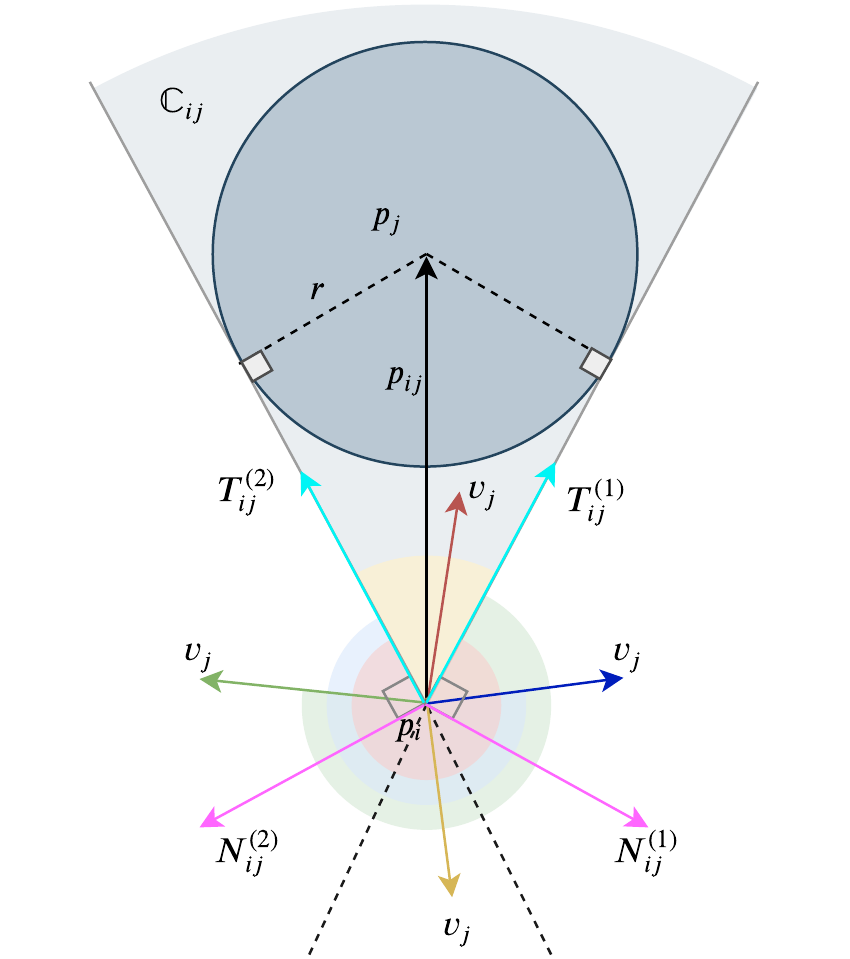}
	\caption{Feasible region analysis for $v_i^k$ according to $v_j^k$}
	\label{fig:vo_simp}
\end{figure}

Based on the velocity obstacle $\mathbb V_{ij}^k$, if the collision avoidance condition $v_i^k\cap \mathbb V_{ij}^k = \emptyset$ is satisfied, the feasible region of $v_i^k$ should be the orange region shown in Fig.~\ref{fig:vo_ori}, according to the direction of $v_j^k$ shown as the blue arrow. Obviously, different directions of $v_j^k$ can result in different feasible regions of $v_i^k$. Therefore, we need to discuss it in terms of different cases, as shown in Fig.~\ref{fig:vo_simp}.

\begin{itemize}
	\item Case 1: $a>0, b>0$\\
	Here, $v_j^k$ is inside the collision cone $\mathbb C_{ij}^k$, which is shown as the red arrow in Fig.~\ref{fig:vo_simp}. The feasible region of $v_i^k$ is shown as the red region, which means 
	\begin{IEEEeqnarray}{rCl}
		v_i^k = c T_{ij,1}^{k} + d T_{ij,2}^{k}, \quad \neg(c>0, d>0),
	\end{IEEEeqnarray}
	where $\neg$ represents the logical negation operation.
	\item Case 2: $a\le0,b\le0$\\
	In this case, $v_j^k$ is inside the inverted cone of $\mathbb C_{ij}^k$, which is shown as the yellow arrow in Fig.~\ref{fig:vo_simp}. The feasible region of $v_i^k$ is in the yellow region, which means 
	\begin{IEEEeqnarray}{rCl}
		v_i^k = c T_{ij,1}^{k} + d T_{ij,2}^{k}, \quad (c>0, d>0).
	\end{IEEEeqnarray}

	\item Case 3: $a<0, b>0$\\
	In this case, $v_j^k$ is shown as the green arrow in Fig.~\ref{fig:vo_simp}, so we have
	\begin{IEEEeqnarray}{rCl}
		v_i^k = cT_{ij,1}^{k} + ev_j^k, \quad  \neg(c>0, e>0).
	\end{IEEEeqnarray}

	\item Case 4: $a>0, b<0$\\
	Here,  $v_j^k$ is shown as the blue arrow in Fig.~\ref{fig:vo_simp}.
	\begin{IEEEeqnarray}{rCl}
		v_i^k = dT_{ij,2}^{k} + ev_j^k, \quad  \neg(d>0, e>0).
	\end{IEEEeqnarray}
\end{itemize}

To summarize all the scenarios as discussed above, let $v_j^k = a T_{ij,1}^{k} + b T_{ij,2}^{k}$, where $a,b \in \mathbb R$, and assume $v_i^k = c T_{ij,1}^{k} + d T_{ij,2}^{k} + ev_j^k$, then we have the following condition:
\begin{IEEEeqnarray}{rCl}
\begin{cases}
e=0,\neg(c>0, d>0), \quad \text{if } a>0, b>0 \\
e=0,\phantom\neg (c>0, d>0), \quad \text{if }  a\leq0, b\leq0 \\
d=0,\neg(c>0, e>0), \quad  \text{if }  a<0, b>0 \\
c=0,\neg(d>0, e>0), \quad \text{if }   a>0, b<0. \\
\end{cases}
\end{IEEEeqnarray}
This condition is a case-by-case discussion about $v_i^k$ according to the direction of $v_j^k$ and there are logical negation operations in the condition, whereby it is sectionally discontinuous and disjunctive, which is hard to compute.

\subsection{Analysis on Collision Cones}
\begin{figure}[thpb]
	\centering
	\includegraphics[width=0.35\textwidth, trim=80 100 50 0,clip]{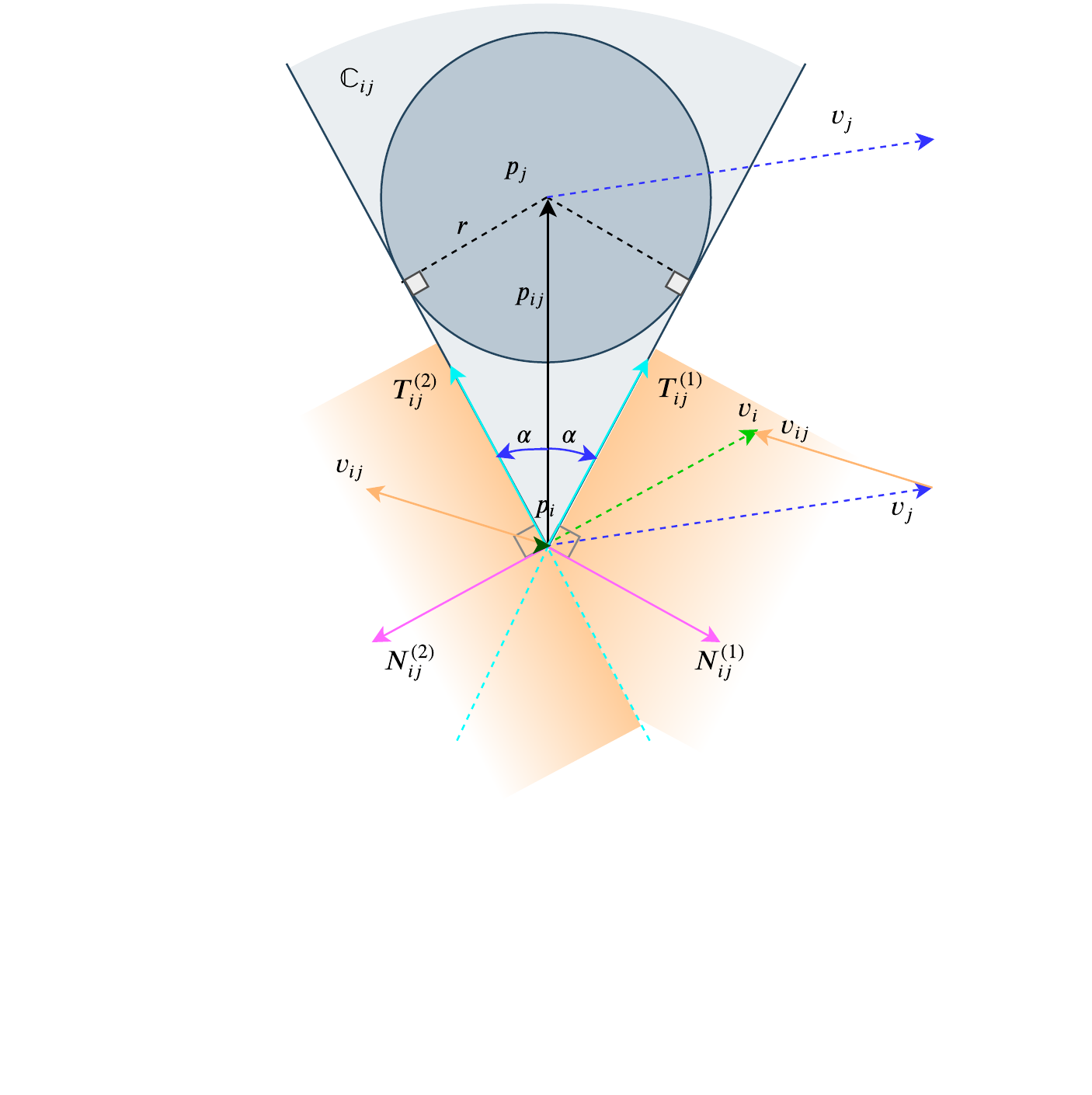}
	\caption{Collision avoidance conditions analysis for $v_{ij}^k$}
	\label{fig:cc}
\end{figure}

In this part, we focus on the relationship between the relative velocity $v_{ij}^k$ and the collision cone $\mathbb C_{ij}^k$, as shown in Fig.~\ref{fig:cc}. In order to meet the collision-free requirement, i.e., there is no intersection between the relative velocity $v_{ij}^k$ and the collision cone $\mathbb C_{ij}^k$, $v_{ij}^k$ should point outside the collision cone, which means that the angle between the two outer normal vectors $N_{ij,1}^k, N_{ij,2}^k$, and $v_{ij}^k$ should be within the range $(-\frac{\pi}{2}, \frac{\pi}{2})$, as shown in Fig.~\ref{fig:cc}.

Two outer normal vectors $N_{ij,1}^{k}, N_{ij,2}^{k}$ are given by
\begin{IEEEeqnarray*}{rCl}
	N_{ij,1}^{k} & = &  \begin{bmatrix} \cos \frac{\pi}{2} & \sin \frac{\pi}{2} \\ -\sin \frac{\pi}{2} & \cos \frac{\pi}{2} \end{bmatrix} T_{ij,1}^{k} \\
	N_{ij,2}^{k} & = &  \begin{bmatrix} \cos \frac{\pi}{2} & -\sin \frac{\pi}{2} \\ \sin \frac{\pi}{2} & \cos \frac{\pi}{2} \end{bmatrix} T_{ij,2}^{k}. \yesnumber
\end{IEEEeqnarray*}
The probabilistic condition of collision avoidance~\eqref{eq:condition} can be transformed as 
\begin{IEEEeqnarray}{rCl}
\left(v_{ij}^k \cdot N_{ij,1}^{k} >0\right) \bigcup\left(v_{ij}^k \cdot N_{ij,2}^{k} >0\right),\; \forall j\in\mathbb N_n, j\neq i.
\end{IEEEeqnarray}
In addition, chance constraints can be rewritten as 
\begin{IEEEeqnarray*}{rCl}
 &\;\;\quad\operatorname{Pr}\left( \left(N_{ij,1}^{k}\cdot v_{ij}^{k} >0 \right)\bigcup \left(N_{ij,2}^{k}\cdot v_{ij}^{k} >0 \right)\right) \ge 1-\delta_{ij} \\
\iff & \operatorname{Pr}\left( \left(N_{ij,1}^{k}\cdot v_{ij}^{k} \leq 0\right) \bigcup \left(N_{ij,2}^{k}\cdot v_{ij}^{k} \leq 0 \right)\right) \le \delta_{ij} \\
\iff & \quad\;\operatorname{Pr}\left( N_{ij,1}^{k}\cdot v_{ij}^k \leq 0 \right) \le \delta_{ij,1}, \\
	 & \quad \operatorname{Pr}\left( N_{ij,2}^{k}\cdot v_{ij}^k \leq 0 \right) \le \delta_{ij,2} \\
\iff & \quad\;\operatorname{Pr}\left( N_{ij,1}^{kT}v_{i}^k \leq N_{ij,1}^{kT}v_j^{k} \right) \le \delta_{ij,1}, \\
	& \quad \operatorname{Pr}\left( N_{ij,2}^{kT} v_{i}^k \leq N_{ij,2}^{kT}v_j^{k} \right) \le \delta_{ij,2}.  \yesnumber
\end{IEEEeqnarray*}

Due to the existence of noise, $v_{ij}^k$ is subject to a normal distribution, then we have 
\begin{IEEEeqnarray}{rCl}
\delta_{ij,1} \delta_{ij,2} = \delta_{ij}.
\end{IEEEeqnarray}

In order to compute probabilistic chance constraints in a deterministic manner, we introduce the following lemma.
\begin{lemma} 
	\label{lemma4}
	Given any matrix $A$ and scalar $b$, for a multivariate random variable $ X(t)$ corresponding to the mean $\mu(t)$ and covariance $\Sigma(t)$, the chance constraint
	\begin{IEEEeqnarray*}{rCl}
	\label{eq:chance_constraint}
	{\operatorname{Pr}\left({A}^T{X}(t) < b \right)}  \leq \varphi,   \yesnumber
	\end{IEEEeqnarray*}
	is equivalent to a deterministic linear constraint
	\begin{IEEEeqnarray*}{rCl}
	\label{eq:chance_constraint_2}
	{{A}^T \mu(t) - b} \ge \eta,  \yesnumber
	\end{IEEEeqnarray*}
	where $\eta = {\sqrt{2{A}^T \Sigma(t) {A}}\operatorname{erf}^{-1}(1-2\varphi)}$,  $\operatorname{erf}$ is the error function $\operatorname{erf}(x) = \frac{2}{\sqrt{\pi}} \int_0^x \exp(-t^2) dt$, and $\varphi$ is the predefined allowable probability threshold of collision. 
\end{lemma}

Notice that the variable $v_i^k$ is a multivariate Gaussian random variable $v_i^k\sim N(\hat v_i^k, W_i)$, we can obtain the equivalent constraint:
\begin{IEEEeqnarray*}{rrCl}
& \operatorname{Pr}\left( N_{ij,1}^{kT}v_{i}^k \leq N_{ij,1}^{kT}v_j^k \right) &\le& \delta_{ij,1} \\
\iff &  N_{ij,1}^{kT} \hat v_{i}^k - N_{ij,1}^{kT}v_j^k &\geq& \kappa_{ij,1} 
\\
& \operatorname{Pr}\left( N_{ij,2}^{kT}v_{i}^k \leq N_{ij,2}^{kT}v_j^k \right) &\le& \delta_{ij,2} \\
\iff &  N_{ij,2}^{kT} \hat v_{i}^k - N_{ij,2}^{kT}v_j^k &\geq& \kappa_{ij,2},\yesnumber
\label{eq:transform_constr}
\end{IEEEeqnarray*}
where $\kappa_{ij,1}=\sqrt{2 N_{ij,1}^{kT} W_i  N_{ij,1}^{k}} \cdot \operatorname{erf}^{-1}(1-2\delta_{ij,1})$ and $\kappa_{ij,2}=\sqrt{2 N_{ij,2}^{kT} W_i  N_{ij,2}^{k}} \cdot \operatorname{erf}^{-1}(1-2\delta_{ij,2})$. 

The calculations of $\kappa_{ij,1}$ and $\kappa_{ij,2}$ require the knowledge on $W_i$, i.e., the covariance of the $i$th robot at the current time. Moreover, the Kalman filter can be used to estimate the other objects velocity in the prediction horizon. 

\begin{figure*}[]
	\centering
	
	\subfigure[2D results of scenario 1]{
		\begin{minipage}[t]{0.4\linewidth}
			\centering
			\includegraphics[width=0.6\textwidth, trim=30 30 10 10,clip]{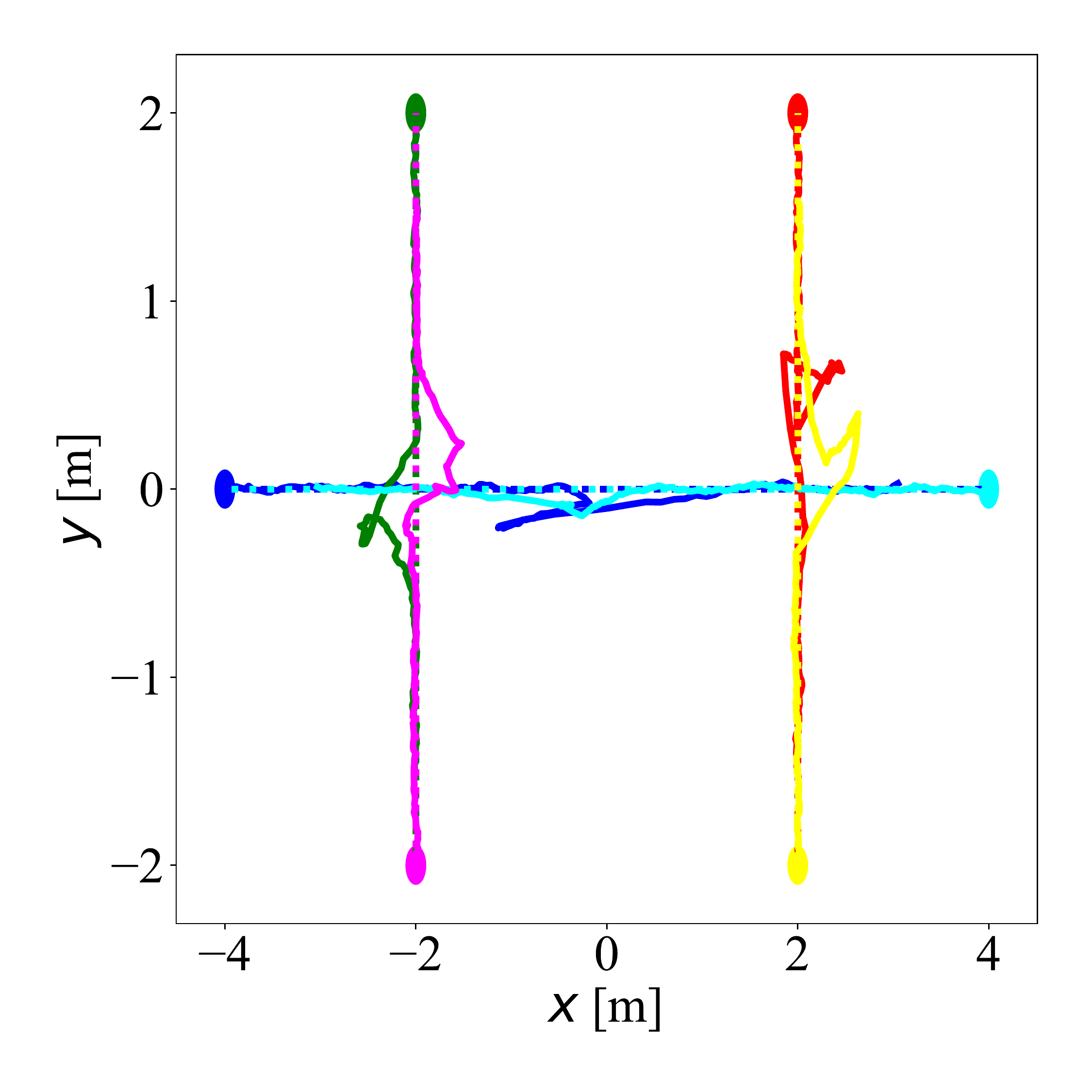}
		\end{minipage}
	}
	\subfigure[3D results of scenario 1]{
		\begin{minipage}[t]{0.5\linewidth}
			\centering
			\includegraphics[width=0.6\textwidth, trim=30 60 30 30,clip]{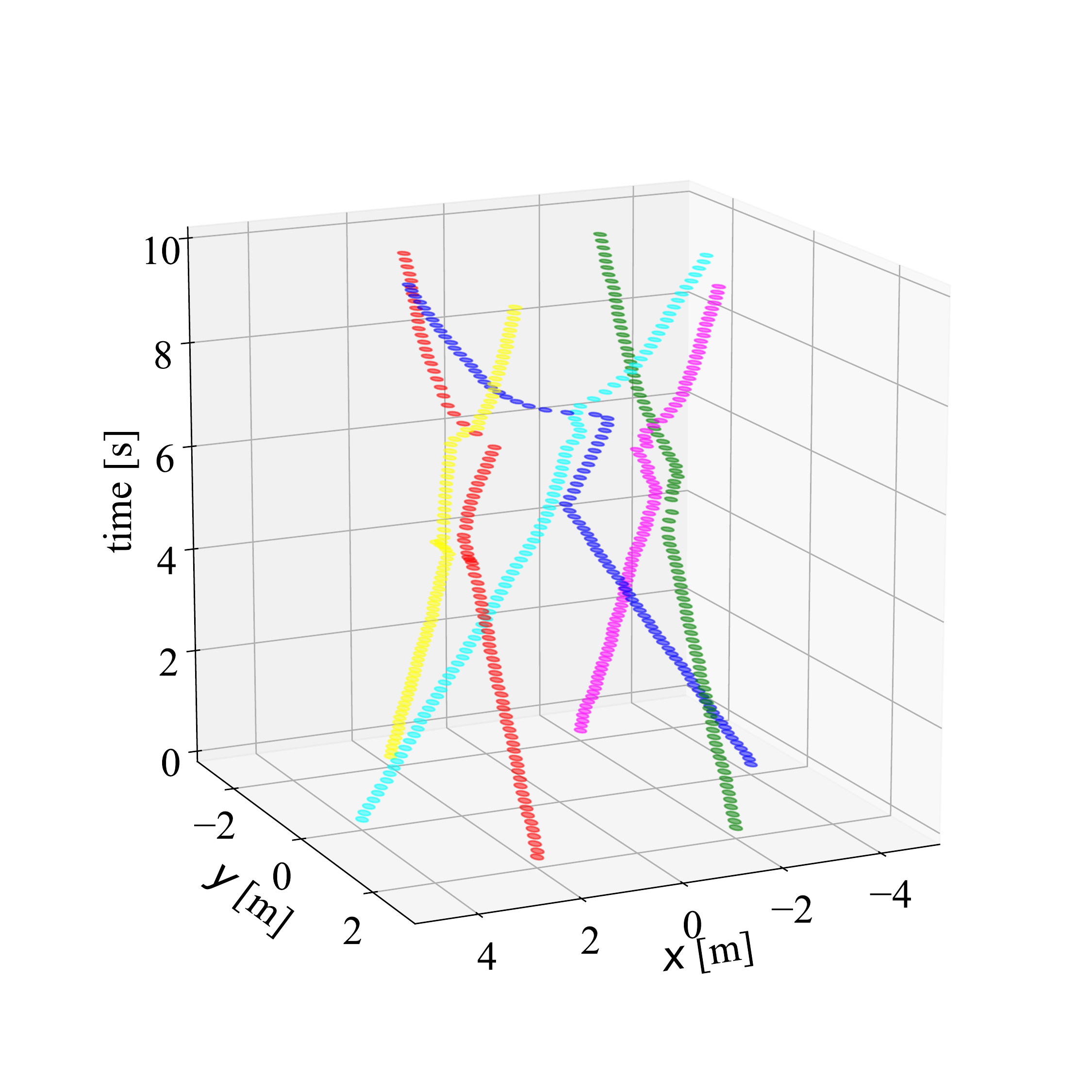} 
		\end{minipage}
	}
	\quad
	\subfigure[2D results of scenario 2]{
		\begin{minipage}[t]{0.4\linewidth}
			\centering
			\includegraphics[width=0.6\textwidth, trim=30 30 10 10,clip]{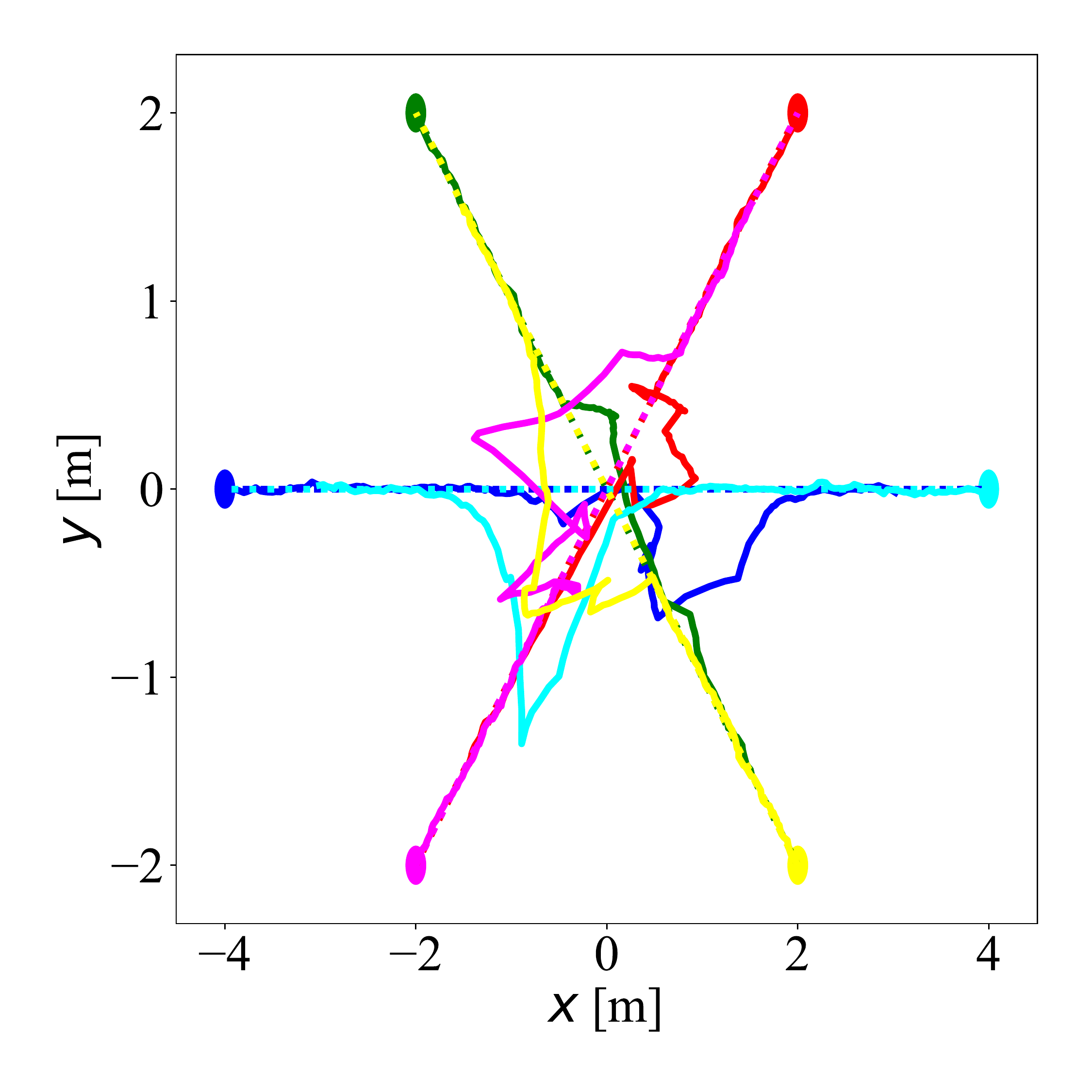}
		\end{minipage}
	}
	\subfigure[3D results of scenario 2]{
		\begin{minipage}[t]{0.5\linewidth}
			\centering
			\includegraphics[width=0.6\textwidth, trim=30 60 30 30,clip]{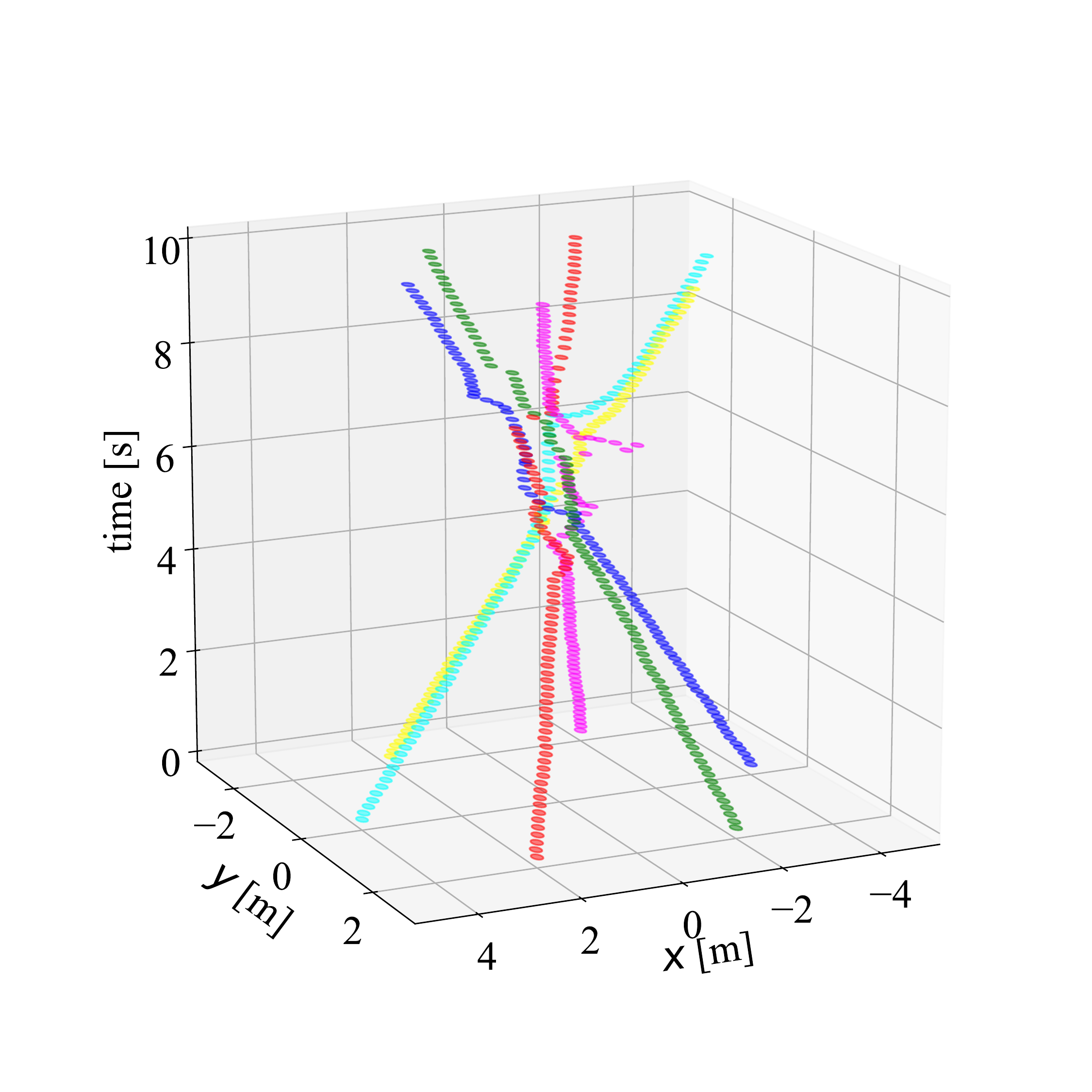} 
		\end{minipage}
	}
	\quad
	\subfigure[2D results of scenario 3]{
		\begin{minipage}[t]{0.4\linewidth}
			\centering
			\includegraphics[width=0.6\textwidth, trim=30 30 10 10,clip]{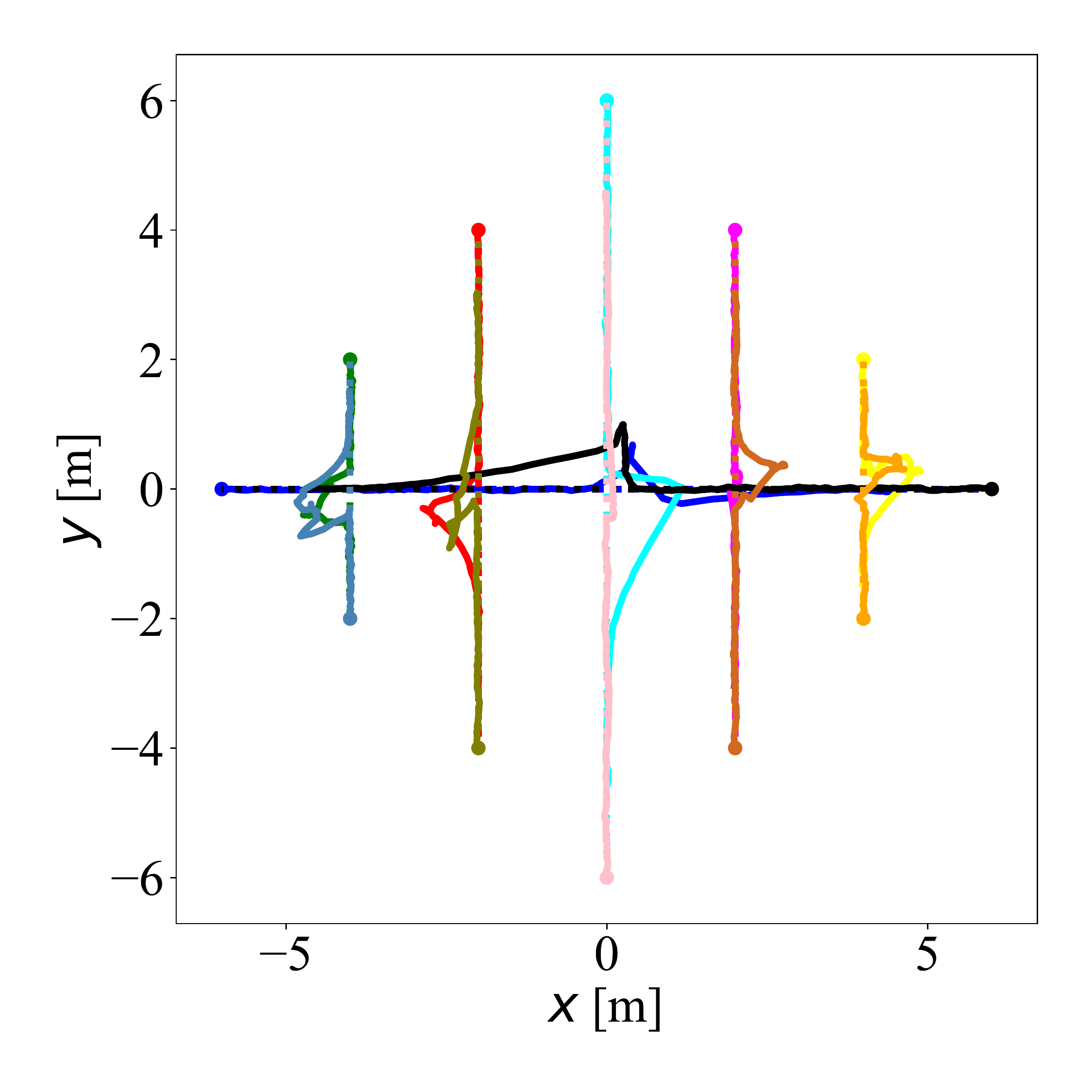}
		\end{minipage}
	}
	\subfigure[3D results of scenario 3]{
		\begin{minipage}[t]{0.5\linewidth}
			\centering
			\includegraphics[width=0.6\textwidth, trim=30 60 30 30,clip]{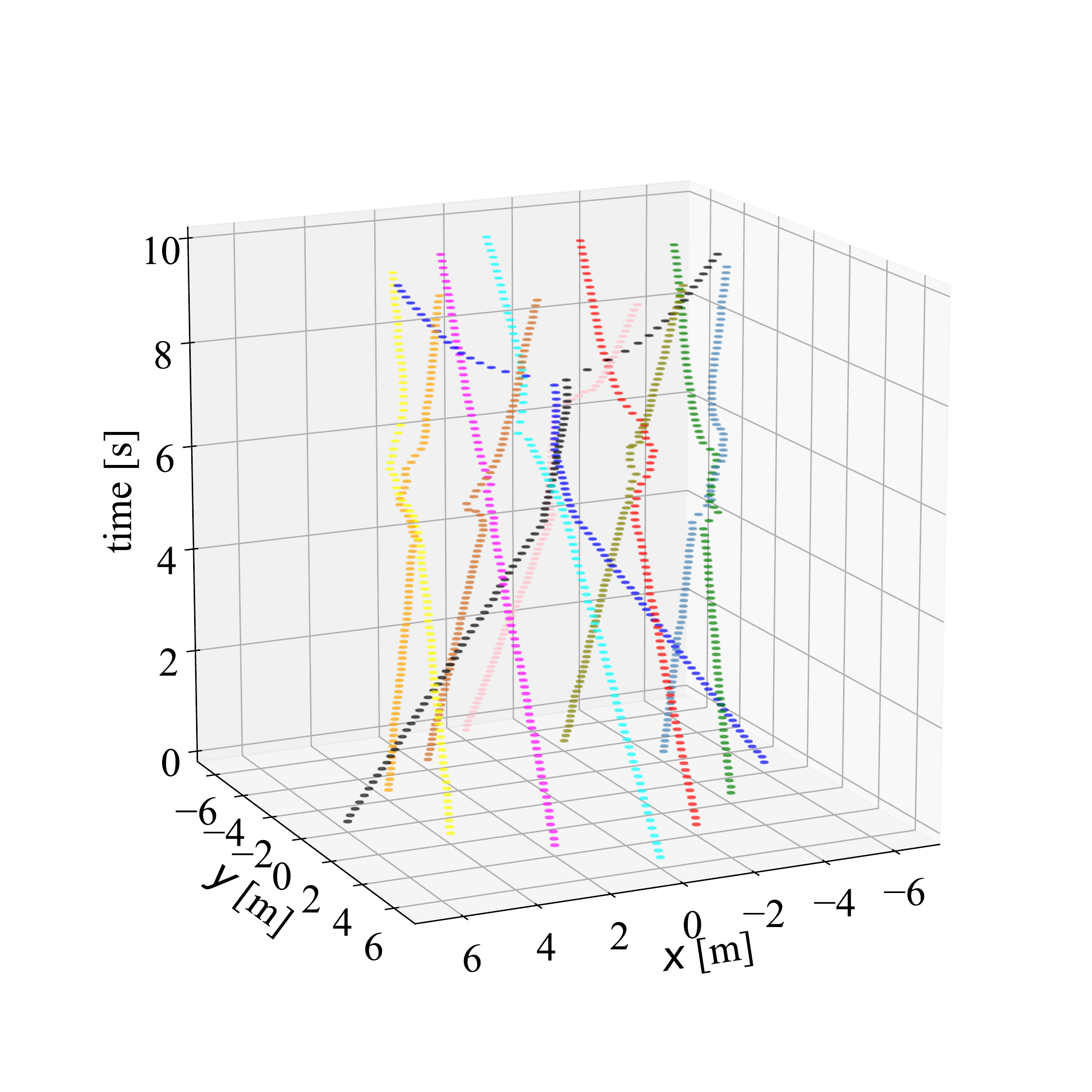} 
		\end{minipage}
	}
	
	\subfigure[2D results of scenario 4]{
		\begin{minipage}[t]{0.4\linewidth}
			\centering
			\includegraphics[width=0.6\textwidth, trim=30 30 10 10,clip]{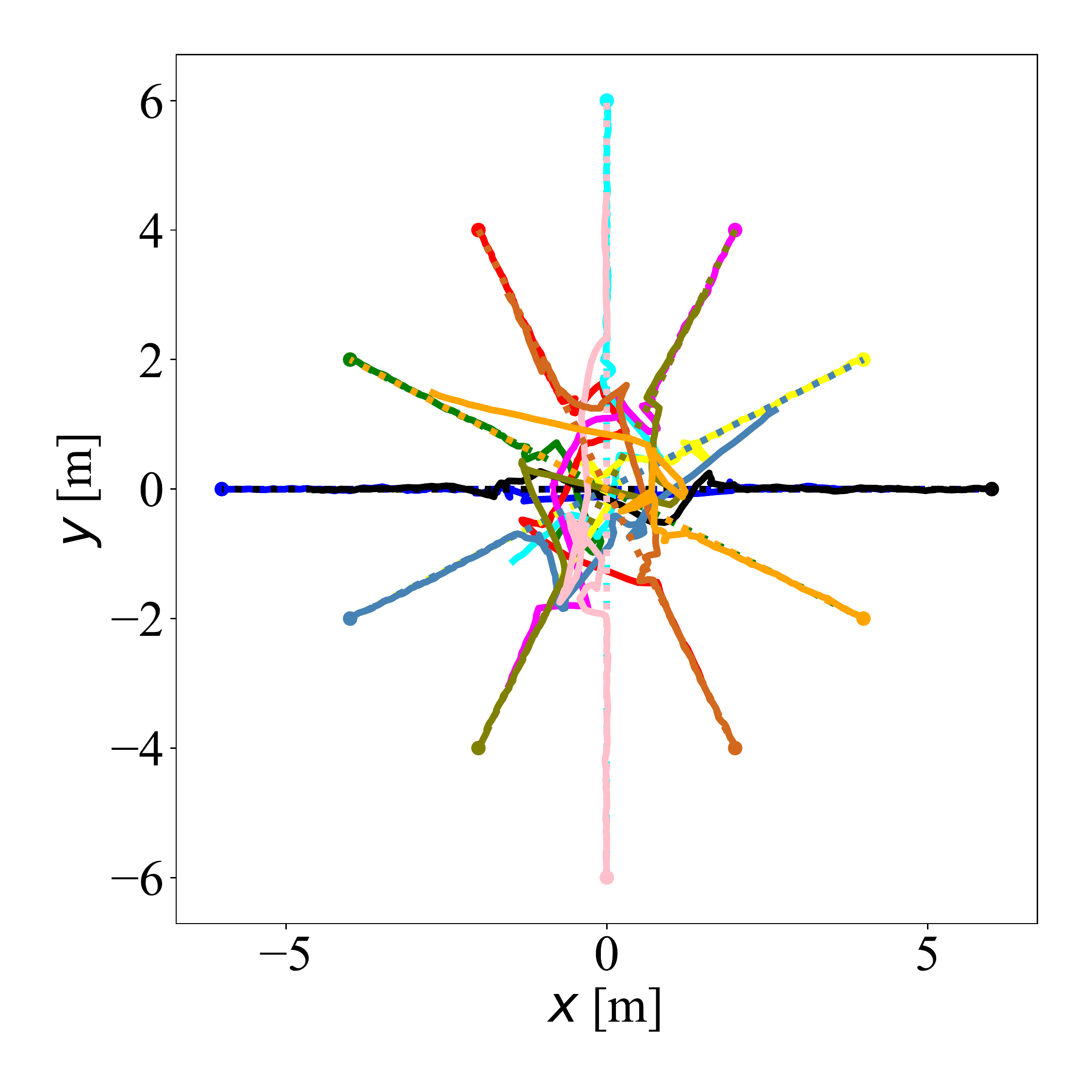}
		\end{minipage}
	}
	\subfigure[3D results of scenario 4]{
		\begin{minipage}[t]{0.5\linewidth}
			\centering
			\includegraphics[width=0.6\textwidth, trim=30 60 30 30,clip]{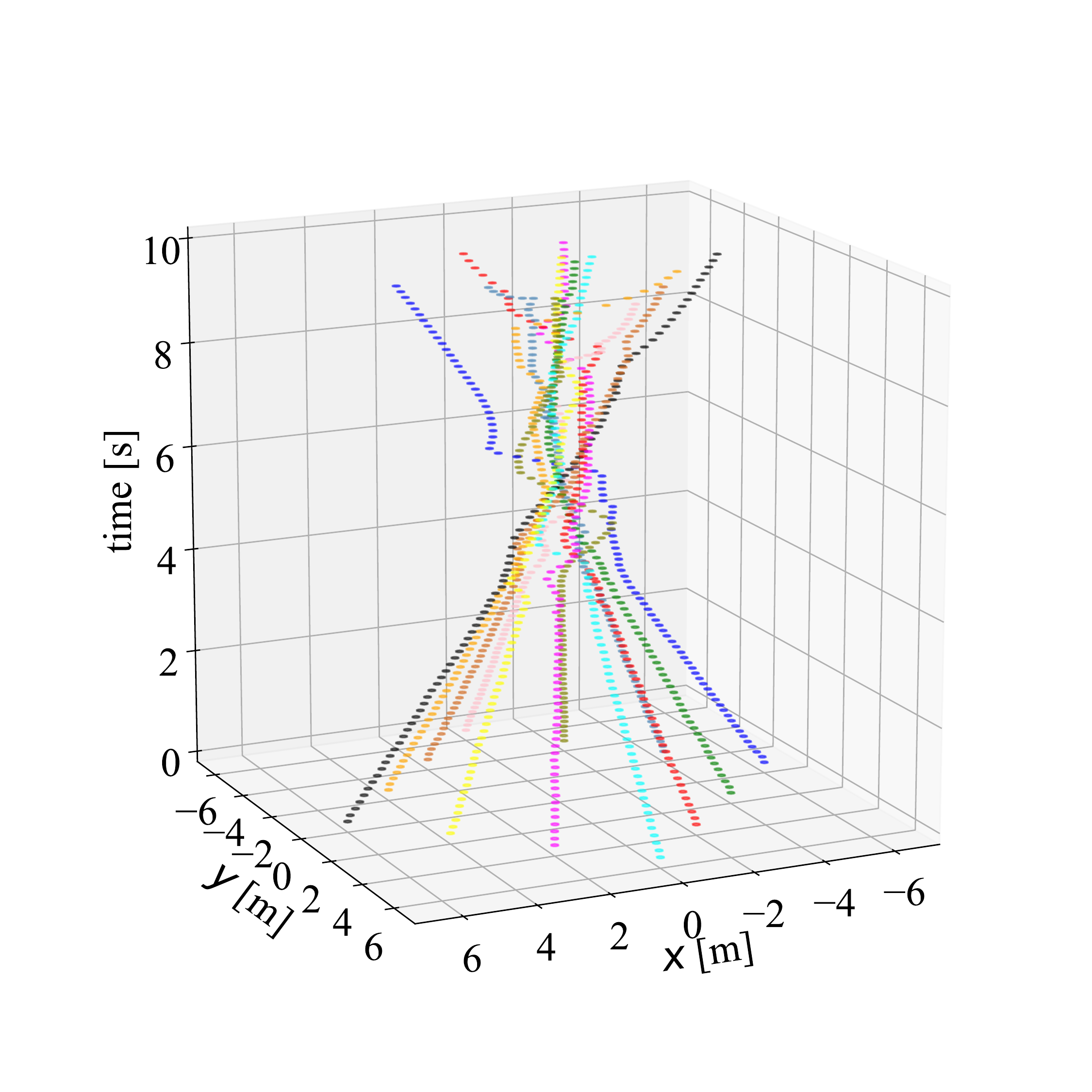} 
		\end{minipage}
	}
	
	\centering
	\caption{Simulation results on multiple objects in different scenarios}
	\label{fig:results}
\end{figure*}

\subsection{Deterministic MPC Formulation and Solving}
\textit{Problem 2: (MPC Problem with Deterministic Chance Constraints)} The probabilistic chance constraints in \eqref{eq:ori_problem} can be transformed into a deterministic manner, which is tractable. This problem can be formulated as
\begin{IEEEeqnarray*}{rCl}
	\min\limits_{x_i^{1:N}, u_i^{0:N-1}} & \quad & \sum_{k=0}^{N-1} \left\|x_i^k-x_{\text{ref},i}^k \right\|_{Q_{i}} + \left\|u_i^k\right\|_{R_{i}} \\
	\operatorname{subject\ to} &\quad& x_i^{k+1} = f_i\left(x_i^k,u_i^k\right), \quad x_i^0 = p \\
	&\quad& N_{ij,1}^{(k+1)T} \hat v_{i}^{k+1} - N_{ij,1}^{(k+1)T}v_j^{k+1} \geq \kappa_{ij,1}, \forall j\in \mathbb N_n, j\neq i \\
	&\quad& N_{ij,2}^{(k+1)T} \hat v_{i}^{k+1} - N_{ij,2}^{(k+1)T}v_j^{k+1} \geq \kappa_{ij,2}, \forall j\in \mathbb N_n, j\neq i \\
	&\quad& \underline x_i^{k+1} \leq x_i^{k+1} \leq \overline x_i^{k+1} \\
	&\quad& \underline u_i^{k} \leq u_i^{k} \leq \overline u_i^{k} \\
	&\quad& v_i^{k+1} \sim \mathcal N\left(\hat v_i^{k+1}, W_i\right) \\
	&\quad& x_i^{k+1}\in \mathcal X_i, \quad u_i^{k} \in \mathcal U_i . \yesnumber
	\label{eq:deterministic problem}
\end{IEEEeqnarray*}
The collision probability between the host object $i$ and all the other objects $j\in \mathbb N_n, j\neq i$ is less than a probability threshold $\delta_i$, where
\begin{IEEEeqnarray}{rCl}
	\delta_i = 1-\prod_{j\in \mathbb N_n, j\neq i}(1-\delta_{ij,1}\delta_{ij,2}).
\end{IEEEeqnarray} 
At time $t$, the cost function is optimized under the constraints in~\eqref{eq:deterministic problem} to obtain the optimal control sequence.
\begin{IEEEeqnarray} {l}
	\label{control_sequence}
	u_{i}^*=\begin{bmatrix} \left(u_i^{*0}\right)^T & \left(u_i^{*1}\right)^T& \cdots & \left(u_i^{*(N-1)}\right)^T \end{bmatrix}^T,
\end{IEEEeqnarray}
and only the first control input $\boldsymbol u_i^{*0}$ will be executed. The pesudocode of how to address the velocity obstacle based receding horizon motion planning with chance constraints is shown in Algorithm~\ref{alg:1}.
\begin{algorithm}
	\caption{Velocity Obstacle Based Receding Horizon Motion Planning with Chance Constraints.}
	\label{alg:1}
	\begin{algorithmic}
		\STATE {\textbf{Initialization:} Dynamic model for all agents; the agents number $n_{num}$; the radius of agents $r_i$; weighting matrices $Q_i$ and $R_i$ in the cost function; initial position and target position of all agents; reference positions of all agents; upper and lower bound of the state $x$ and control input $u$, i.e., $\overline x, \underline x$ and $\overline u, \underline u$, respectively; runtime of MPC $r_{run}$; sampling time $\tau_s$; covariance matrix $W_i$; threshold $\delta_{1}$ and $\delta_{2}$. }
		\STATE {Set the time step $t = 0$.}
		\WHILE {$t<=t_{run}$}
		\FOR {$i\in \mathbb N_n$}
		\FOR {$k=0,1,\cdots, N-1$}
		\STATE {Compute $x_i^{k+1}$ for the $i$th agent based on $x_i^k$ by using~\eqref{eq:dynamics}.}
		\STATE {Generate constraints of physical limitations for the $i$th agent.}
		\STATE {Compute the two transformed deterministic constraints for the $i$th agent by \eqref{eq:transform_constr}.}
		\ENDFOR
		\STATE {Compute the quadratic cost function for the $i$th agent.}
		\STATE {Solve the deterministic optimization problem~\eqref{eq:deterministic problem}.} \STATE {Obtain the optimal control sequence $u_i^*$.}
		\STATE {Execute the first control input $u_i^{*0}$ for the $i$th agent.}
		\STATE {Pass the updated states of neighbors $j\in \mathbb N_n, j\neq i$ to the $i$th agent.}
		\ENDFOR
		\STATE {$t=t+\tau_s$.}
		\ENDWHILE
	\end{algorithmic}
\end{algorithm}


\section{Results}
\label{section:results}
This section describes the implementation of the proposed method, and the effectiveness of the method is evaluated by simulations.  All the relevant parameters of the simulation are shown in Table~\ref{tab:setting_params} in the Appendix. Here, we add the Gaussian noise to the velocity of objects model, and the added measurement noise is zero mean with the covariance $W_i$. Taking the noisy measurements as inputs, a Kalman filter is employed to estimate the state of other moving objects. All of the simulations are implemented in Python 3.7 environment on a PC with Intel i5 CPU@3.30 GHz. The video demonstrating the results can be found at \url{https://www.youtube.com/watch?v=MwA6eUhIAw4}, and the source code is available at \url{https://github.com/Lisnol1/Velocity-Obstacle-Motion-Planning}.

Here, four different scenarios for 6 and 12 objects are designed for validation purposes, and the simulation results are shown in Fig.~\ref{fig:results}. Different colors represent different objects with radius $r_i$. In the simulation, the radius of all objects is set as $r_i=0.2$ m. The circles in (b), (d), (f), and (h) of Fig.~\ref{fig:results} demonstrate the approximate shape of objects. The solid lines and dotted lines in (a), (c), (e), and (g) of Fig.~\ref{fig:results} represent the planned trajectories and reference trajectories for each object, respectively. 
The scenario 1 in Fig.~\ref{fig:results} means that there are 6 objects whose target position is the symmetry point of the initial position along the $x$ or $y$ axis. For example, for the green object, its initial position is (-2, 2) m and then its target position is (-2, -2) m. Fig.~\ref{fig:results}(a) shows the trajectories of the 6 objects and Fig.~\ref{fig:results}(b) presents the position of all objects in each time step. According to Fig.~\ref{fig:results}(a) and (b), there is no collision happening in this scenario, as no two objects occur in the same position at the same time. The scenario 2 in Fig.~\ref{fig:results} indicates that the 6 objects starting from their initial positions need to pass through the origin point (0, 0) m without any collision and reach their target positions that are the symmetry points of the initial positions along with the origin, i.e., along the both $x$ and $y$ axis. To be specific, the green object in scenario 2 starts from (-2,2) m and needs to reach its target position (2, -2) m. The results of the scenario 2 are shown in Fig.~\ref{fig:results}(c) and (d). Also, all objects are able to arrive at their destinations with no collision, based on the two subfigures. Similarly, the task in scenario 3 is to reach the target positions that are symmetry points along the x or y axis of the initial positions for all 12 objects. Fig.~\ref{fig:results}(e) and (f) illustrate the resulted trajectories of these 12 objects, and it is straightforward to observe that all of the 12 objects can reach their target positions without collisions. In scenario 4, there are 12 objects whose target positions are the symmetry points of the initial positions along with the origin, i.e., along the both $x$ and $y$ axis. The trajectories of these 12 objects are shown in Fig.~\ref{fig:results}(g) and (h), and we can observe that all of the trajectories are collision-free, which satisfies our requirement. Overall, all the subfigures from Fig.~\ref{fig:results}(a) to (h) indicate that this approach can effectively avoid collisions with the other moving objects (including agents and obstacles, which can be represented by different dynamics and shapes). 

\begin{figure}[H]
	\centering
	\includegraphics[width=0.4\textwidth, trim=0 0 0 0,clip]{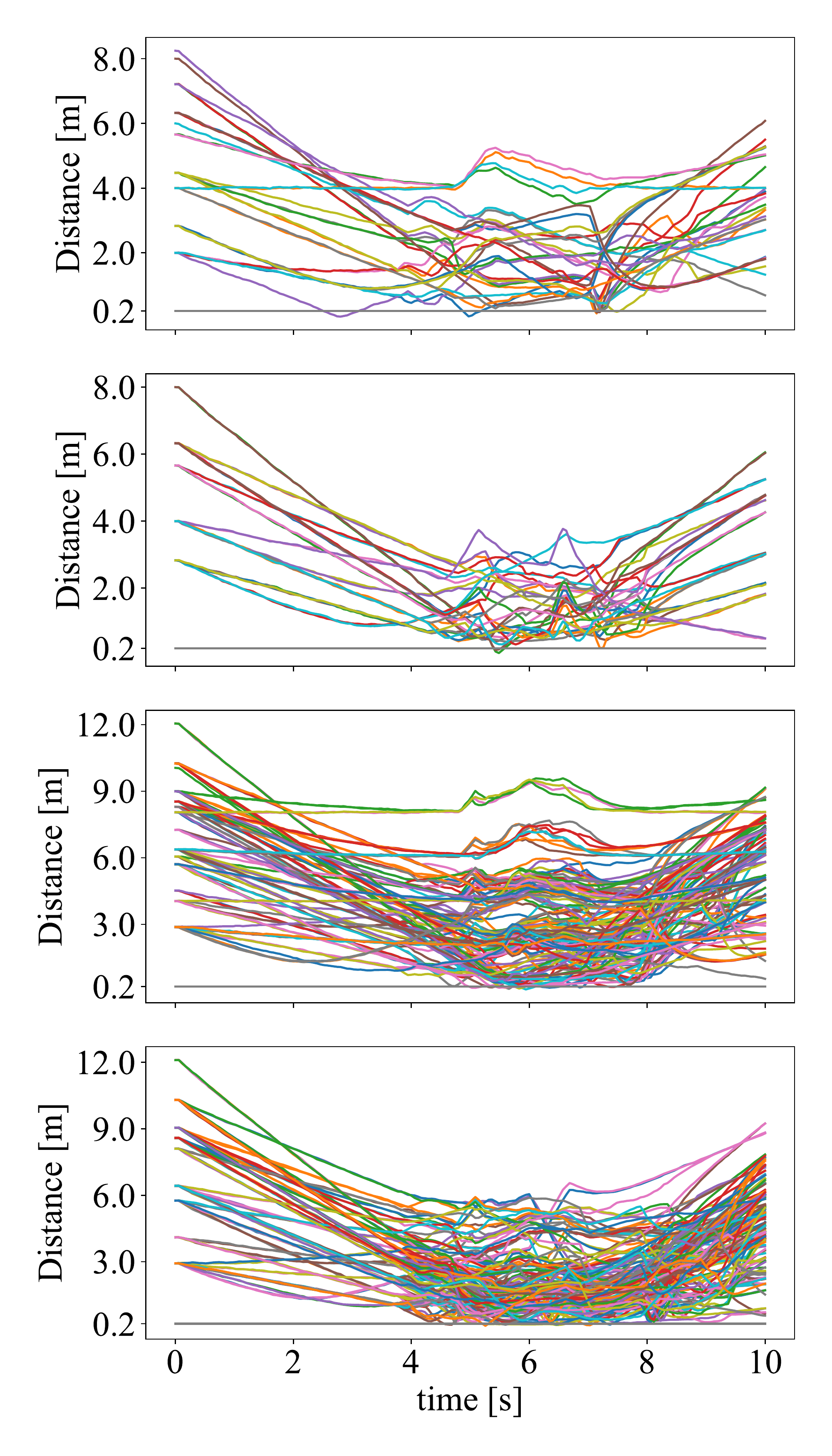}
	\caption{Distance between each pair of objects in all scenarios}
	\label{fig:distance}
\end{figure}

The distances between each pair of objects in all scenarios are shown in Fig.~\ref{fig:distance}. The subfigures from the first one to the fourth one are the results from scenario 1 to scenario 4, respectively. According to all these subfigures in Fig.~\ref{fig:distance}, it is apparent that each object can maintain a certain safety distance from other objects, as the minimum safety distance is greater than 0.1 m, which is the radius of each object. Therefore, there is no collision happening in all of the four scenarios.  

For scenario 4, we compare the results of our method under different three levels of measurement noise ($\frac{1}{4}W_i$, $W_i$, and $4W_i$) with the deterministic MPC method, which is proposed in \cite{nageli2017real}. The minimum distance between each pair of agents and the success rate are treated as the safety metrics. The comparison results are shown in Table~\ref{tab:diff_noise}. Besides, due to a tighter bound of collision probability approximation, our method can keep a larger minimum distance during running under the same noise level, compared with the deterministic MPC method. Also, with a larger noise level, our proposed method maintains the success rate of 100\%, but the success rate of deterministic MPC decreases from 71\% to 41\%. According to Table~\ref{tab:diff_noise}, we can observe that our proposed method achieves higher safety performance compared with the deterministic MPC method.
\begin{table}[H]	
	\caption{Trajectory Safety Comparison of Two Algorithms with Different Levels of Noise (The Values Are Computed from Successful Runs).}
	\label{tab:diff_noise}
	\begin{tabular}{|c|c|c|c|}
		\hline
		Noise &  Safety metrics   & Deterministic MPC~\cite{nageli2017real} & Our method  \\ \hline
		\multirow{2}{*}{$\frac{1}{4}W_i$} & Minimum distance  & 0.112 m   & 0.167 m                      \\ \cline{2-4} 
		& Success rate       & 72 \%   & 100 \%                        \\ \hline
		\multirow{2}{*}{$W_i$}            & Minimum distance    & 0.136 m    & 0.194 m                   \\ \cline{2-4} 
		& Success rate        & 61 \%     & 100 \%                     \\ \hline
		\multirow{2}{*}{$W_i$}            & Minimum distance    & 0.148 m   & 0.231 m                    \\ \cline{2-4} 
		& Success rate         & 41 \%     & 100 \%                   \\ \hline
	\end{tabular}
\end{table}

\begin{figure}[H]
	\centering
	\includegraphics[width=0.4\textwidth, trim=0 0 0 0,clip]{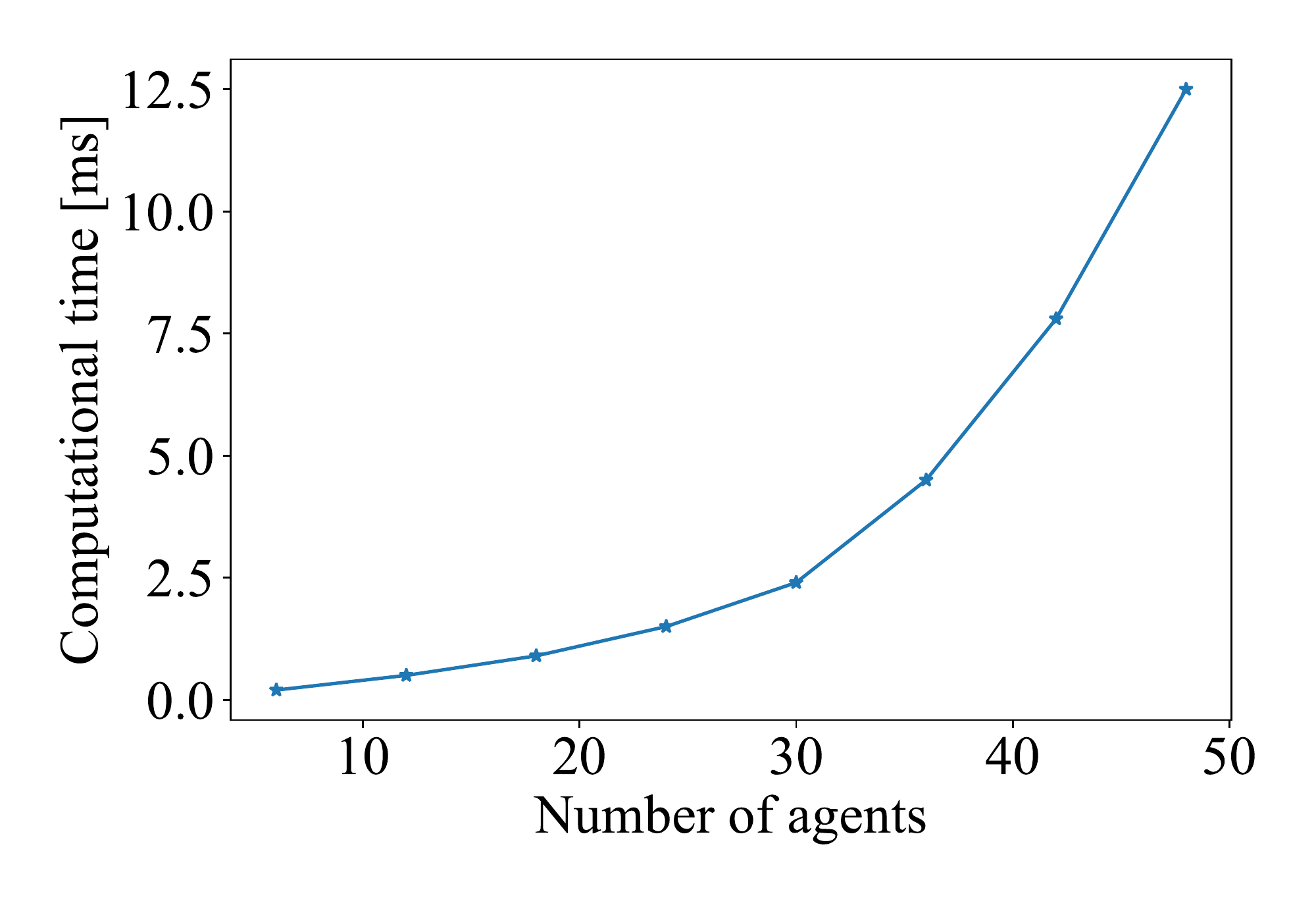}
	\caption{Distance between each pair of objects in all scenarios}
	\label{fig:comp_time}
\end{figure}

Fig.~\ref{fig:comp_time} shows the computational time with an increasing number of agents moving across the origin circle. In this case, the initial positions and the target positions of all agents are symmetric along the origin, i.e., both the $x$ and $y$ axis. Based on Fig.~\ref{fig:comp_time}, the computational time for different agent numbers are less than the sampling time $\tau_s = 0.05\, \textup{s}$, which means that the real-time implementation can be realized. Besides, it is straightforward to observe that the computational time increases with the number of agents increasing. 

\section{Conclusion}
\label{section:conclusion}
In this paper, a velocity obstacle based receding horizon motion planning method, is proposed for potential collision avoidance. Based on the feasible region of $v_i$ provided by the velocity obstacles method, a chance constrained RHC problem is formulated and solved. A feasible region of velocity for the host agent is derived and formulated as probabilistic collision constraints. Hence the proposed method can generate the trajectories at the velocity level. Besides, this method can also provide a probability threshold of potential collision during the motion planning process. Several simulation scenarios for multiple agents are employed to validate the effectiveness and efficiency of our proposed methodology. In terms of the future work, one prospective research is to realize the motion planning in 3-dimensional space. Moreover, another future work is to consider the uncertainty in shape and velocity of the agents at the same time.

\section*{APPENDIX}
\label{appendix:param}
Table~\ref{tab:setting_params} shows the values of parameters in Section~\ref{section:vo_ccProblem}.
\begin{table}[h]
	\renewcommand{\arraystretch}{1.5}
	\caption{Parameter Setting}
	\label{tab:setting_params}
	\begin{center}
		\begin{tabular}{|p{.13\textwidth}<{\centering}|p{.05\textwidth}<{\centering}|p{.18\textwidth}<{\centering}|p{.02\textwidth}<{\centering}| }
			\hline
			Meaning & Notation & Value &Unit \\  \hline
			Radius & $r_i$ & 0.1 & m \\ \hline
			Mass & $m$ & 1 & kg \\ \hline
			Covariance of noise & $W_i$ & diag(0.01, 0.01, 0.05, 0.05) & -  \\ \hline
			Threshold & $\delta_{ij,1}, \delta_{ij,2}$ & 0.1, 0.1 & -  \\ \hline
			Sampling time & $\Delta t$ & 0.05 & s \\ \hline
			Prediction horizon & $N$ & 25 & -  \\ \hline
			State weighting matrix & $Q_i$ & $\operatorname{diag}(10, 10, 1, 1)$ & -  \\ \hline
			Control input weighting matrix & $R_i$ & $\operatorname{diag}(1, 1)$ & -  \\ \hline
			Maximum/minimum state limitations & $\overline x_i/ \underline x_i$ & $[\infty \  \infty \ 10 \ 10]^T$ / $[-\infty \  -\infty \ -10 \ -10]^T$   & -\\ \hline
			Maximum/minimum input limitations & $\overline u_i/ \underline u_i$ & $[\infty \  \infty]^T$ / $[-\infty \  -\infty]^T$   & -  \\ \hline
		\end{tabular}
	\end{center}
\end{table}

\bibliographystyle{IEEEtran}
\bibliography{IEEEabrv,Reference}

\end{document}